\documentclass[conference]{IEEEtran}

\IEEEoverridecommandlockouts
\pdfoutput=1
\usepackage{hyperref}

\usepackage{cite}
\usepackage{svg}
\usepackage{amsmath,amssymb,amsfonts}
\usepackage{cleveref}
\usepackage{algorithmic}
\usepackage{subfigure}
\usepackage{graphicx}
\usepackage{textcomp}
\usepackage{xcolor}
\usepackage[numbers,sort]{natbib}
\def\BibTeX{{\rm B\kern-.05em{\sc i\kern-.025em b}\kern-.08em
    T\kern-.1667em\lower.7ex\hbox{E}\kern-.125emX}}
\begin{document}

\IEEEoverridecommandlockouts
\IEEEpubid{\makebox[\columnwidth]{ 979-8-3503-5067-8/24/\$31.00~\copyright2024 IEEE \hfill} 
\hspace{\columnsep}\makebox[\columnwidth]{ }}
\IEEEpubidadjcol

\title{Training Interactive Agent in Large FPS Game Map with Rule-enhanced Reinforcement Learning\\

}

\author{
    \IEEEauthorblockN{Chen Zhang$^{1,2}$, Huan Hu$^{2}$, Yuan Zhou$^{2}$, Qiyang Cao$^{2}$, Ruochen Liu$^{2}$, Wenya Wei$^{2}$, Elvis S. Liu$^{2,*}$\thanks{*Corresponding author}}
    \IEEEauthorblockA{$^1$ School of Software Engineering, University of Science and Technology of China, Hefei, China}
    \IEEEauthorblockA{$^2$ Tencent Games}
    \IEEEauthorblockA{zhangchenzc@mail.ustc.edu.cn,
    \\\{luckyhu, ariellezhou, hughyycao, ruochenliu, wenyawei, elvissyliu\}@tencent.com}
}

\maketitle

\begin{abstract}
 In the realm of competitive gaming, 3D first-person shooter (FPS) games have gained immense popularity, prompting the development of game AI systems to enhance gameplay. However, deploying game AI in practical scenarios still poses challenges, particularly in large-scale and complex FPS games. In this paper, we focus on the practical deployment of game AI in the online multiplayer competitive 3D FPS game called Arena Breakout, developed by Tencent Games. We propose a novel gaming AI system named Private Military Company Agent (PMCA), which is interactable within a large game map and engages in combat with players while utilizing tactical advantages provided by the surrounding terrain.

To address the challenges of navigation and combat in modern 3D FPS games, we introduce a method that combines navigation mesh (Navmesh) and shooting-rule with deep reinforcement learning (NSRL). The integration of Navmesh enhances the agent's global navigation capabilities while shooting behavior is controlled using rule-based methods to ensure controllability. NSRL employs a DRL model to predict when to enable the navigation mesh, resulting in a diverse range of behaviors for the game AI. Customized rewards for human-like behaviors are also employed to align PMCA's behavior with that of human players.
\end{abstract}

\begin{IEEEkeywords}
game AI, deep reinforcement learning, navigation mesh, shooting rules, self-play, rule-enhanced.
\end{IEEEkeywords}
\section{Introduction}
First-person shooter (FPS) games in 3D have gained immense popularity in the competitive gaming realm. As these games have evolved from early titles like Maze War and Half-Life to more recent ones such as Apex Legends, CS: GO, and Valorant, there has been a growing interest in developing intelligent AI systems for FPS games. Traditional decision-making approaches based on behavior trees (BT) have proven inadequate in exploring all possible decisions within complex 3D environments. To address this limitation, deep reinforcement learning (DRL) has been introduced as a flexible alternative for designing game AI in 3D FPS games.

However, despite significant advancements that have shown the power of DRL agents~\citep{wu2016training,lample2017playing,jaderberg2019human, pearce2022counter,chen2022wild}, the practical deployment of game AI in FPS games still faces challenges. Existing environments used for training and testing AI agents often have small game maps and limited game duration, which do not align with the scale and complexity of modern FPS games. For example, in environments like VizDoom~\citep{wu2016training}, the duration of each match is typically around 100 frames. Additionally, the complexity of 3D FPS game environments, with physically modeled terrain and objects, poses difficulties for environment perception and global navigation for AI agents. Furthermore, balancing global navigation and combat engagement within the game AI presents significant demands, requiring simultaneous macro-level navigation and micro-level combat decision-making. There is also a need to ensure that the game AI exhibits both competitive strength and human-like behavior.
\begin{figure}
    \centering
    \includegraphics[width=\linewidth]{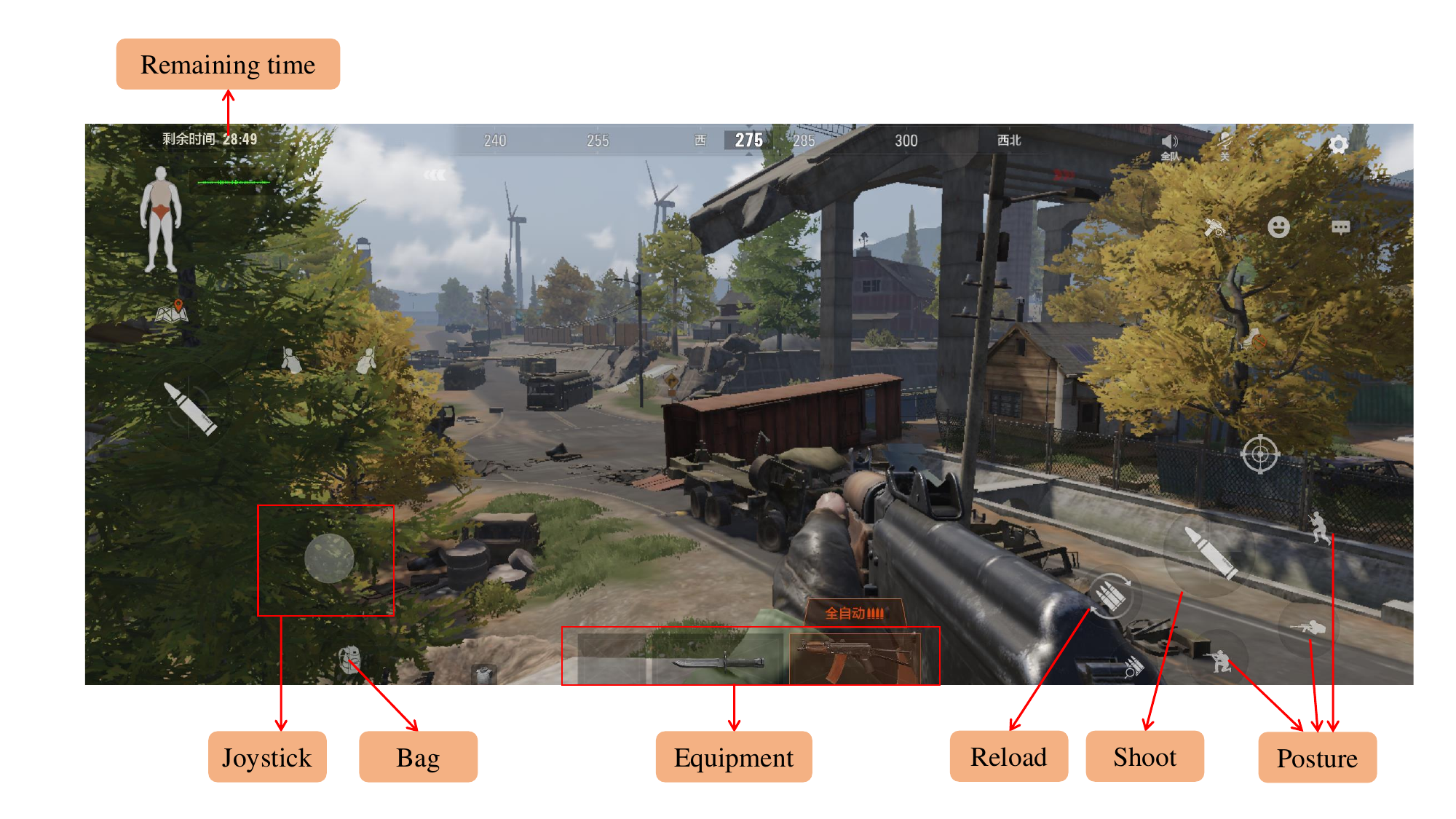}
    \caption{The interface of Arena Breakout}
    \label{fig:interfance}
\vskip -0.2in
\end{figure}

In this paper, we try to solve these issues and focus on the practical deployment of game AI in an online multiplayer 3D FPS game developed by Tencent Games called Arena Breakout. Players in Arena Breakout control a first-person perspective game character and aim to reach a designated location for evacuation within a limited timeframe. Throughout the game, players interact with other players or game AI, either by evading or eliminating them, to achieve evacuation. 

To enhance players' gaming experience, we propose a novel game AI system named Private Military Company Agent (PMCA) which is interactiable in a large game map. PMCA is primarily deployed in matches involving high-level players. When PMCA encounters players at any location on the map, it will initiate attacks and engage in combat with the player, utilizing the surrounding terrain to gain tactical advantages. In order to achieve the objectives, PMCA must not only address the challenges of navigation and combat in modern 3D FPS games but also perform effectively in large-scale matches with an average duration exceeding forty minutes. Furthermore, PMCA needs to constrain its behavior so that does not confuse the players.

To tackle these challenges, we propose a novel method that combines Navigation Mesh (Navmesh)~\citep{ref} and shooting rules with reinforcement learning (NSRL). The integration of Navmesh enhances the agent's global navigation capabilities while considering firefights and shooting behavior is executed using rule-based methods to enhance controllability. Unlike previous approaches that directly incorporate rules into the program, NSRL takes a more subtle approach by using a DRL model to predict whether to enable the navigation mesh. The decision to switch to the navigation mesh is made only when the DRL model determines it is appropriate, resulting in a more diverse range of behaviors for the game AI. Additionally, customized rewards for human-like behaviors are employed to ensure that PMCA's behavior aligns with human players, further enhancing its human-like behavior.

Overall, we construct the Markov decision process (MDP) of Arena Breakout and employ the Proximal Policy Optimization (PPO) \cite{schulman2017proximal} algorithm to update the policy. Experimental results show the ability of PMCA in global navigation and the diversity of behavior.

The contributions of this paper are as follows:
\begin{itemize}
    \item We propose a novel game AI system based on rule-enhanced reinforcement learning, integrating Navmesh and shooting rules into deep reinforcement learning (DRL) to enhance the performance of the DRL agent. This design allows the agent to balance global navigation across the entire map while accurately aiming and firing at targets.
    \item We deploy the system in Arena Breakout and engage in long-term interactions with players, representing a significant milestone for the practical application of DRL.
\end{itemize}
\begin{figure*}[htb]
    \centering
    \includegraphics[width=\textwidth]{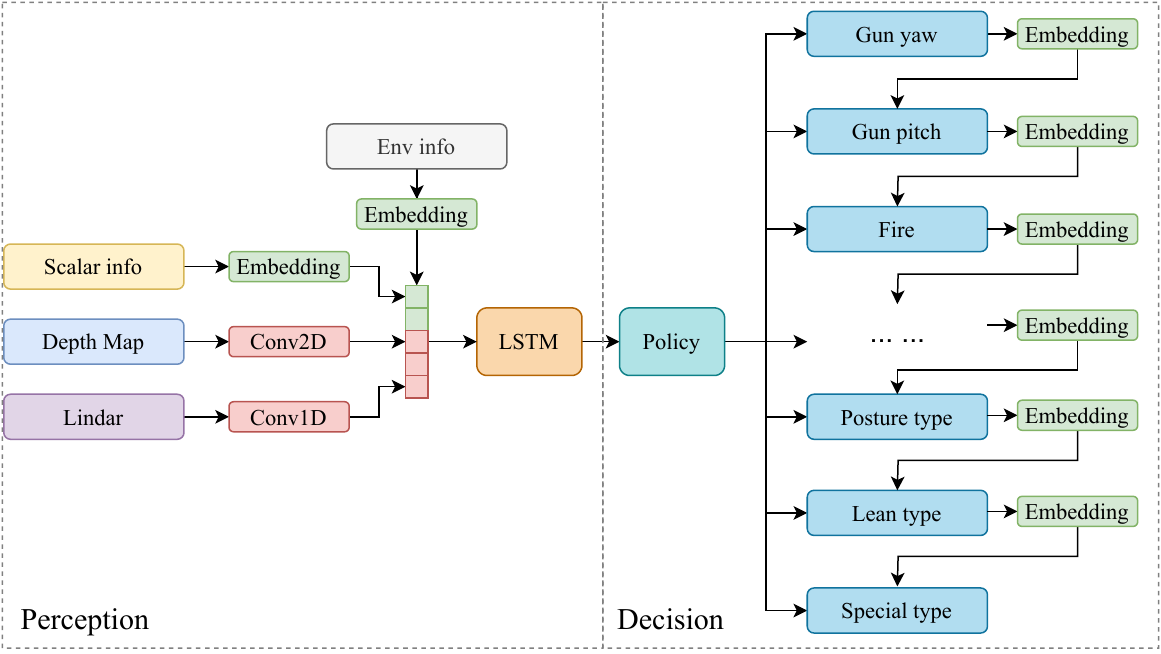}
    \caption{Framework of PMCA.}
    \label{figframe}
\vskip -0.2in
\end{figure*}
\section{Notation And Background}
\subsection{Arena Breakout}

Arena Breakout is a 3D first-person shoot game developed by Tencent Games based on Unreal Engine 4. Its interface is shown in \cref{fig:interfance}. At the time of writing this paper, it has more than 80 million registered players worldwide. Different from the games mentioned above, Arena Breakout not only retains the original shooting game logic but also adds more tactical content. Players need to compete for resources and reach a designated location for evacuation within a limited timeframe. The objective of the game is to survive while obtaining as many resources (weapons, equipment, etc.) as possible, storing them in a secure box,
and reaching the evacuation point to complete the evacuation. 
Each player has a secure box, and resources stored in the box will not be lost upon death, while items outside the box will be lost. Therefore, the core of the gameplay is how players can obtain the maximum amount of resources while staying alive. Owing to developed by Unreal Engine 4, Arena Breakout features complex 3D modeling and detailed maps where most objects are physically modeled. The player in Arena Breakout requires intricate operation, as players need to control their direction and movement while aiming and firing at targets. The game also includes actions such as crouching, crawling, and jumping to change the player's posture. The complex environment and diverse operations add more replayability to the game while increasing the difficulty of training the game AI.

\subsection{State Space and Action Space}\label{mdp}
In order to apply reinforcement learning algorithms to Arena Breakout, we need to construct the Markov Decision Process (MDP) for the game. We decompose Arena Breakout into the state space and the action space.

In the state space of the Arena Breakout, we have devised multiple information sources concerning the agent, encompassing scalar game variables and perception information derived from raycast detection. The scalar game variables, including the agent's position, rotation angle, orientation, and combat status (such as health and engagement in combat), are directly acquired through the APIs provided by Unreal Engine 4. These variables are subjected to mathematical processing and synthesizing them into five distinct categories of fundamental information for input into the neural network (Table \ref{table1}). Each category of information influences the agent's decision-making and behavior differently.
The perception information based on raycast detection is also implemented through the Unreal Engine 4. The agent emits rays from its position towards entities within the game and captures pertinent details, such as the entity's surface normal vector. Consequently, the agent can ascertain the distance and position of the entity relative to itself utilizing these rays. 
We further process this information to generate depth maps and encapsulate the circular ray, which are subsequently employed as inputs for the neural network. 

The operation of Arena Breakout is achieved through a joystick that determines the direction combined with different action buttons. Therefore, in the action space, we imitate the left and right-hand operations of human players. Based on Euler rotation angles, we divide the executable actions into nine action heads (\cref{table1}). Each action head is represented by a one-hot vector that indicates the execution state of that action. Fire determines whether to initiate the firing action. Gun\_yaw and Gun\_pitch represent how to adjust the position of the gun barrel. Move\_type determines the movement mode, such as running or walking. Path\_type indicates the pathfinding method, where the model determines whether to use atomic movement or Navmesh movement. Motion represents the movement direction, indicating a source motion at a fixed angle. Posture\_type expresses the current character's posture, such as crouching, prone, jumping, etc. Lean\_type indicates whether to perform a leaning action. Special\_type represents special operations in the game, such as aiming down sights (ADS) or other specific actions.

\begin{table}[]
\vskip 0.1in
\centering
\begin{tabular}{|c|c|c|c|}
\hline
State    & Type  & Dim         \\ \hline
Basic\_info    & float & {[}1,124{]}               \\
Opponent\_info   & float & {[}1,99{]}               \\
Env\_info  & float & {[}1,301{]}                           \\
Depth Map      & float & {[}40,80{]}             \\
Lindar\cite{baker2019emergent}         & float & {[}144,3{]}              \\ \hline
Action         & Type & Dim       \\ \hline
Fire           & int  & {[}2{]}               \\
Gun\_yaw  & int  & {[}13{]}              \\
Gun\_pitch & int  & {[}8{]}               \\
Move\_type   & int  & {[}4{]}              \\
Path\_type & int  & {[}4{]}              \\
Motion     & int  & {[}17{]}               \\
Posture\_type  & int  & {[}4{]}               \\
Lean\_type     & int  & {[}3{]}               \\
Special\_type  &int &{[}3{]} \\ \hline
\end{tabular}
\caption{State space and action space of Arena Breakout}\label{table1}
\vskip -0.2in
\end{table}

\subsection{Reinforcement Learning}

In a standard reinforcement learning setting, the agent interacts with the environment and receives full perceptual information about the state. The Markov decision process (MDP) of Arena Breakout is formulated as a 5-tuple 
$\langle \mathcal{S},\mathcal{A},\mathcal{P},r,\gamma\rangle$,
where $\mathcal{S}$ and $\mathcal{A}$ are the state and action spaces; 
$r: \mathcal{S} \times \mathcal{A} \to \mathbb{R}$ and $\mathcal{P}: \mathcal{S} \times \mathcal{A} \to \Delta_{\mathcal{A}}$ are the reward function and transition probability distribution, with $r(s, a) \in [R_{min}, R_{max}]$ and $P(\cdot|s, a)$ being the reward and the next state probability of taking action a in state s; $\gamma$ is discount factor.
In each step t, the agent gets a state $s_{t} \in \mathcal{S}$ as the current state of the environment. The agent predicting an action $a_{t} \in \mathcal{A}$ given the state $s$ with the policy $\pi(a_{t}|s_{t})$. Agent gets a reward $r(s_{t}, a_{t})$ after the action is executed. The state value function $V_{\pi}(s)=\mathbb{E_{\pi}}[\sum\limits_{k=0}^{T}\gamma^{k}R_{t+k+1}|S_t=s]$ and the action value function $Q_{\pi}(s)=\mathbb{E_{\pi}}[\sum\limits_{k=0}^{T}\gamma^{k}R_{t+k+1}|S_t=s, A_t=a]$ are always leveraged to optimize policy.
The advantage function $A(s,a)=Q(s,a)-V(s)$ is used to measure the quality of action $a$ compared to the average quality.

The goal of the RL algorithm is to find an optimal policy $\pi^*$ to maximize the expectation of the discounted cumulative reward:
\begin{equation}
    \pi^{*} = \arg\max\limits_{\pi}\mathbb{E}[\sum\limits_{t=0}^{n}\gamma^{t}r(s_{t}, a_{t})]
\end{equation}

Several DRL algorithms have been proposed, some use value-based methods to find an optimal value function such as Deep Q-net work (DQN)\cite{hosu2016playing}, and others use policy gradient to optimize policy with the gradient of reward\cite{silver2014deterministic}. 
\citet{schulman2017proximal} utilizing importance sampling to improve convergence efficiency, the characteristic of data reuse also enables the application of distributed algorithms. Our work is based on PPO.

\section{METHODS}
Modern 3D FPS shooter games typically exhibit characteristics such as complex environments and large-scale maps. The need for precise positioning in specific locations poses challenges for global navigation. In Arena Breakout, PMCA is required to pursue any encountered player on the map, further emphasizing the need for accurate navigation. On the other hand, engaging in combat with players presents challenges as modern 3D FPS games often simulate firearm recoil, resulting in realistic bullet trajectories. This becomes a training challenge for the agent, as it must learn how to control the bullet trajectory to hit opponents accurately. This section will discuss how to train a globally interactive game on a $1000 \times 400$ (unit: meters in UE4) map and the approaches adopted to address these two challenges. The results of these approaches will be demonstrated in the \cref{exper}.

\subsection{Framework}
We illustrate our framework as shown in \cref{figframe}. In each step, the game's basic information is separately input into a feature extraction module. All scalar information undergoes feature extraction through a feed-forward network. The depth map and the Lindar~\cite{baker2019emergent} is input into the feature extraction module. By utilizing two-dimensional convolution to perceive the contour features of objects represented in the depth map and employing one-dimensional circular convolution to extract surrounding terrain features, we achieve a multi-modal fusion-based environmental perception approach by fusing the features from each feature extraction module with the environment information of the game after concatenation.

To give the model memory capacity, we utilized LSTM (Long Short-Term Memory) networks. The features filtered through the LSTM are then input into the policy network and value network, which are used for action prediction and value calculation respectively. As mentioned in \cref{mdp}, we divided the action space into nine action heads based on human operation. Each action head is represented by a one-hot vector that indicates the execution state of the corresponding action. There are certain dependencies among the action heads. For example, according to the limitations of the game itself, the model should not predict both firing and movement simultaneously. To decouple the dependencies among actions, we did not adopt the approach of parallel prediction for action heads. Instead, we sequentially output the action heads using a hierarchical action mask with auto-regressive embedding \cite{vinyals2019grandmaster}.

In typical reinforcement learning tasks, action masking is employed to block corresponding actions in specific states to enhance exploration efficiency.  
In our policy network, all action heads are sequentially generated. Except for the first action head, the output of each subsequent action head is determined jointly by the embedding of the previous action and the policy network's output. Through auto-regressive embedding, the policy implicitly learns the dependencies between preceding and succeeding actions and propagates them layer by layer. During action prediction, the policy can temporarily mask actions in subsequent layers that conflict with the current action. The policy also determines whether to mask the actions in the current layer based on the preceding actions. In this way, we conduct hierarchical action masking through auto-regressive prediction, ensuring action compliance while improving exploration efficiency. After all action heads have completed their outputs, these actions are delivered to the game client for execution.

\subsection{Navigation Mesh and Shooting-rule Enhanced Reinforcement Learning}
To address the challenges of global navigation on a $1000 \times 400$ (unit: meters in UE4) map and the issue of firing when encountering enemies in-game, we employ Navigation Mesh And Shooting-rule enhanced Reinforcement Learning (NSRL). This approach combines the integration of a Navigation Mesh (Navmesh) and atomic shooting rules to enhance the performance of the game's AI.

\textbf{Global Navigation.} In the context of global navigation, simply incorporating Navmesh directly into the program would lead to rule-based behavior in game AI, thereby sacrificing the diversity provided by DRL models. Therefore, in NSRL, a more gentle approach is employed. The decision-making power to use the Navmesh is delegated to the DRL model, allowing the model to predict whether to enable the Navmesh. By setting Navmesh as a predictable action, NSRL maintains the ability for global navigation while still preserving the diversity of the game AI provided by the DRL model.
\begin{figure}[ht]
    \centering
    \includegraphics[width=\linewidth]{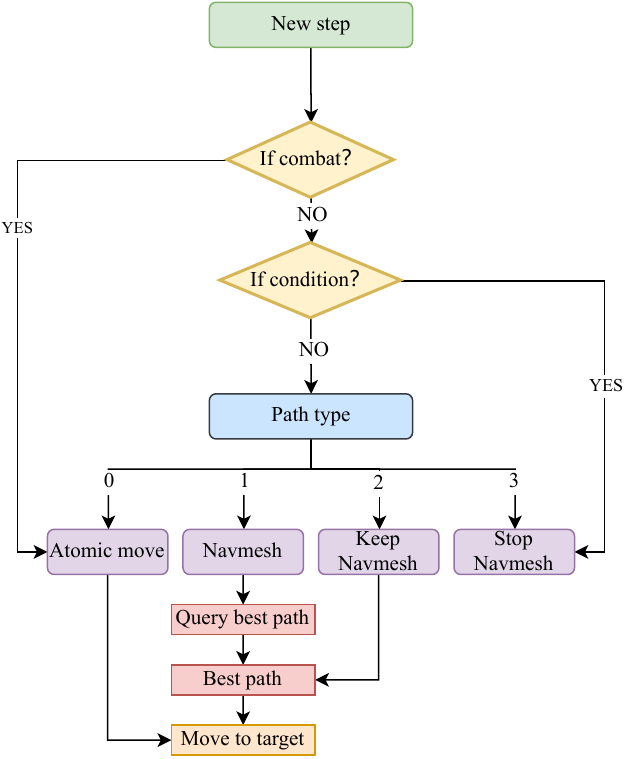}
    \caption{The illustration of Navmesh enhanced global navigation.}
    \label{fig:Navmesh}
\vskip -0.1in
\end{figure}

\begin{figure*}[htb]
    \centering
    \includegraphics[width=\linewidth]{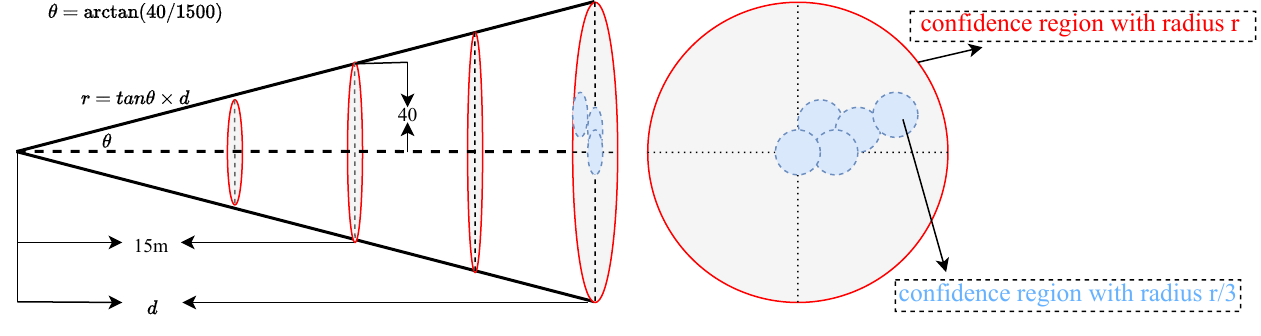}
    \caption{The illustration of shoot rules.}
    \label{fig:shoot rules}
\vskip -0.2in
\end{figure*}

As \cref{fig:Navmesh} shown, in each step, the DRL agent will predict the Path type before motion. When the DRL model predicts atomic movement, it also needs to select an orientation from a set of 16 orientations, with each orientation spaced at a fixed angle interval. Subsequently, the agent moves forward a fixed distance according to the specified orientation. When the model chooses to use the Navmesh (\cref{fig:Navmesh}), it searches for the optimal path between the agent and the target using the Navmesh and moves along that path. The Path type includes an action to keep the previous Navmesh, which aims to avoid re-calling the Navmesh when the target is stationary, reducing resource consumption.

It is not feasible to call the Navmesh throughout the entire game as it would be indistinguishable from directly using Navmesh for movement. To address this, we divide the entire game into time slices. At fixed intervals, the client will check for any requests to use the Navmesh. Once a request to use the Navmesh is detected, the predictions of the DRL model for other Path types are masked, and the agent executes the Navmesh until it reaches the target, encounters an enemy, or reaches the time limit for Navmesh usage. This approach ensures global navigation while reducing resource consumption and preserving the diversity inherent in DRL models. Note that when the agent is in a combat scenario, it typically implies that the opponent is visible to the agent. Therefore, to ensure smooth combat, atomic movement is preferred over Navmesh usage. Navmesh is usually predicted when pursuing enemy targets.

\textbf{Shooting Rules.} In Arena Breakout, where realistic firearm ballistics and recoil are simulated, NSRL only needs to predict the gun barrel direction and whether to fire to ensure shooting control. This approach offers the advantage of maintaining stable shooting while enhancing the diversity of shooting actions. In combat scenarios, the baseline firing point is determined based on the visibility priority of the opponent's body parts. A shooting confidence region is calculated that defines the area within which random shots are fired, while shots outside the confidence region are truncated. As shown in \cref{fig:shoot rules}, We have set $\arctan(40/1500)$ as the baseline angle, and the confidence region radius $r$ for different distances is calculated based on the baseline radius. Additionally, we have incorporated an extra confidence region between two consecutive shots, which is represented by a circle with a radius of $r/3$. This imitates the behavior of human players controlling recoil in the game, simulating the concept of "burst shooting" where shots are fired within a certain range to maintain accuracy. 

\subsection{Reward Design}
In reward design, it is important to ensure that the agent has a comprehensive understanding of both global navigation and shooting. Therefore, when designing the rewards, it is necessary to consider global navigation, shooting performance, and the final reward. Since the PMCA is primarily used in high-level matches, it needs to exhibit a certain level of strength. Given a horizon $T$, the final reward $r_T$ can be defined as follows:
\begin{equation*}
    r_T = \left\{
\begin{split}
    &20, if\ win, \\
    -&20, if\ lose, \\
    -&25, if\ draw,
\end{split}
    \right.
\end{equation*}
The navigation reward $r_d$ is $r_d = \Delta_d \times 0.05$. Where $\Delta_d$ is the distance between agent and opponent.
We have implemented additional auxiliary rewards $r_{aux}$ for the agent's shooting behavior to enhance its human-like characteristics. These rewards are categorized into two parts: combat and movement, based on expert priors. This ensures that our agent does not exhibit confusing behaviors that might perplex players. The final reward r defined as:
\begin{equation*}
    r = r_T + r_d + r_{aux}
\end{equation*}

\subsection{Training Process}
\begin{figure}
    \centering
    \includegraphics[width=\linewidth]{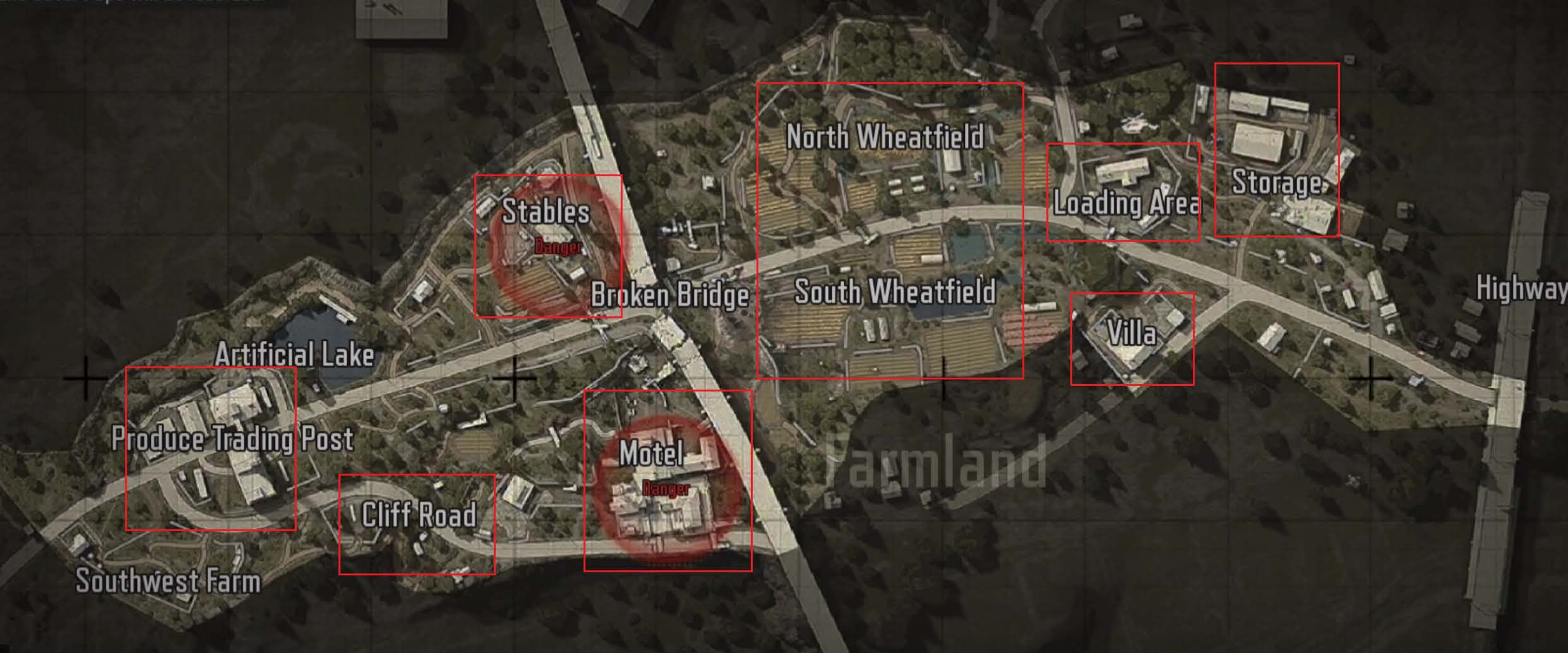}
    \caption{The map in Arena Breakout, the farmland.}
    \label{fig:map}
\vskip -0.2in
\end{figure}

To train the agent on a large $1000 \times 400$ (unit: meters in UE4) map, we initially divided the entire map into eight regions (red square in \cref{fig:map}). At the beginning of each match, the agent and the opponent are randomly generated within the same region. There is guaranteed to be cover or obstacles between the spawn points of the two sides. When the two sides encounter each other, a battle begins as long as one side has a line of sight to the other. This approach not only trains the agent's combat abilities but also enhances its strategic decision-making by encouraging the use of non-combat tactics to defeat opponents. During the training process, an episode ends when either one of the players dies or the time reaches the timeout limit.

PPO is used in our system for policy improvement. We employed the importance-weighted actor-learner Architecture (IMPALA) \cite{espeholt2018impala} algorithm as the primary algorithm in our training framework, which enables agents to learn from multiple sources of experience. The agent interacts with the environment to collect data. After several episodes, the data obtained by the agent is used to estimate the advantage function. We use Generalized Advantage Estimation\cite{schulman2015high} to estimate the advantage function at each time step t:
\begin{equation}
    \hat{A}=\sum\limits_{t=0}^{T}{(\gamma\lambda)}^{t}A_{t+1}^{\pi_{\theta}}
\end{equation}
And the advantage function $A$ is defined as:
\begin{equation}
    {A}^{\pi_{\theta}}(s_{t}, a_{t})=r(s_t, a_t)+\gamma V^{\pi_{\theta}}(S_{t+1})-V^{\pi_{\theta}}(S_{t})
\end{equation}
Where $a_{t}\sim\pi_{\theta}(\cdot|s_{t})$. $r(s_{t},a_{t})$ is the reward function according to the current state and action. $V^{\pi_{\theta}}$ is the value function, which defines the state value of the current policy.

After having the advantage function, we can calculate the loss of the policy. We use PPO with clip, and the loss function is defined as:
\begin{equation}
    \mathcal{L}^{\pi}(\theta)=\min[\rho(\theta)\hat{A}_{t}, clip(\rho(\theta),1-\epsilon,1+\epsilon)\hat{A}],
\end{equation}
to constrain the distance between the sample policy and the target policy. The value loss is defined as:
\begin{equation}
    \mathcal{L}^{v}(\phi)=\mathbb{E}_{a\sim\pi_{\theta}}[\sum\limits_{k=t}^{T}\gamma^{k-t}r(s_{k}, a_{k})-V(s_t)]^2
\end{equation}
And the total loss of current policy $\pi_\theta$ is defined as:
\begin{equation}
    \mathcal{L}(\theta)=\mathcal{L}^{\pi}(\theta)-\alpha\mathcal{L}^{v}(\phi)+\beta\mathcal{L}^{e}
\end{equation}

$\mathcal{L}^{e}$ is the entropy of policy which enhances the exploration. 
\section{Experiment}\label{exper}
\subsection{Experimental Setup}
In this section, we will present our experimental results. We will demonstrate the improvements brought by NSRL in global navigation, shooting, and final win rate. The experiments were conducted using 8 GPUs and 3200 CPU cores, with 4 actors, 4 learners, and 6000 clients involved in each training session. All experiments were based on the PPO algorithm. To ensure fairness, all training sessions used the same set of resources.

\begin{figure*}[ht]
\vskip -0.1in
	\centering
	\subfigure[Visualization of RL Global Navigation] {\includegraphics[width=.49\textwidth]{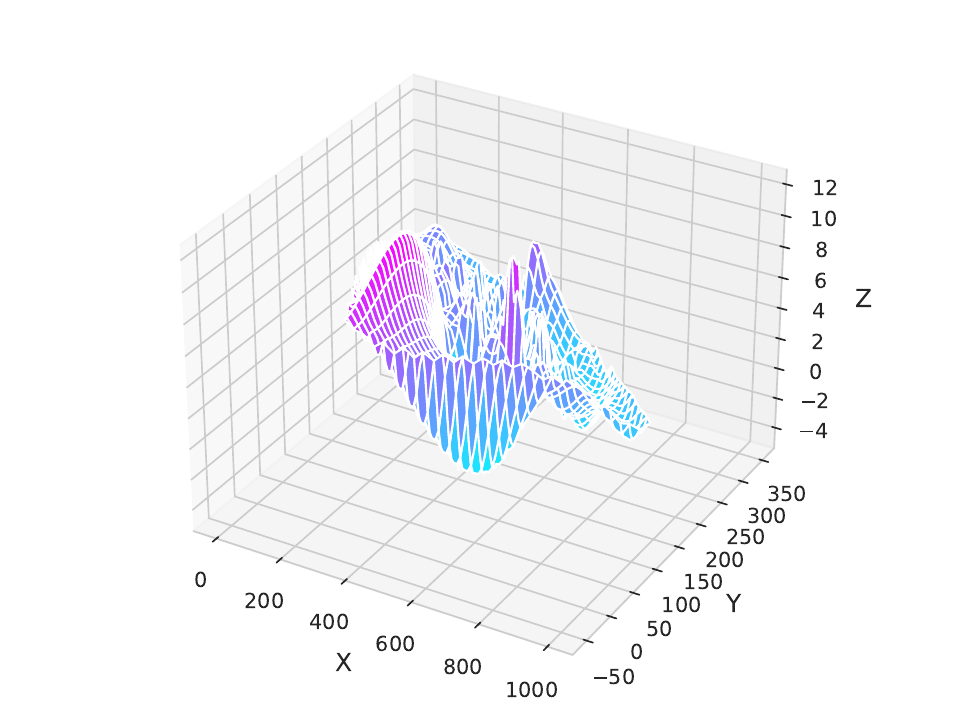}\label{RL}}
	\subfigure[Visualization of NSRL Global Navigation] {\includegraphics[width=.49\textwidth]{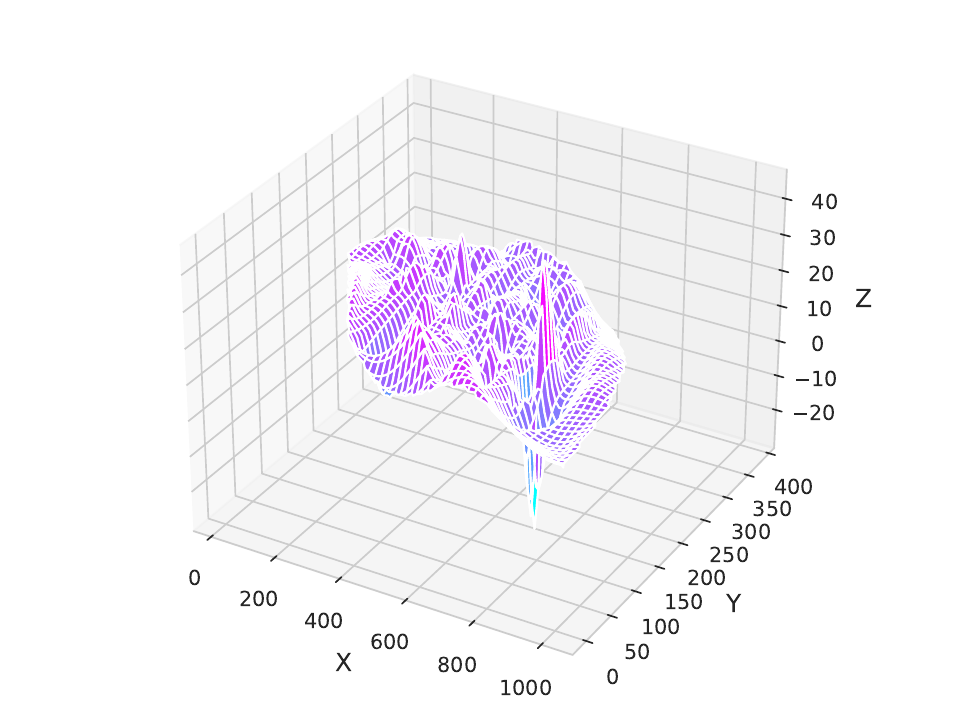}\label{NSRL}}
	\caption{Visualization of RL and NSRL Global Navigation. The closer the color is to purple, the higher the number of traversals, while the closer it is to blue, the fewer the number of traversals. The NSRL visualization shows that most areas are closer to purple, indicating a higher number of traversals, while there are fewer areas close to blue. On the other hand, the RL visualization reveals that only a small portion is close to purple, suggesting limitations in RL's global navigation capabilities.}
	\label{visual_nav}
\vskip -0.2in
\end{figure*}

\subsection{Results}
\textbf{Global Navigation.} To compare the global navigation performance of the NSRL agent and the RL agent under identical conditions, we conducted a study where both agents were trained and tested in the same environment. The NSRL and RL agents were trained using the same expert prior behavior tree as their opponent. The starting points and in-game resources were kept consistent across the experiments. In this scenario, we recorded the traversal points of both the NSRL agent and the RL agent on the map. We collected data from over 1000 game sessions for analysis. The NSRL and RL agent global navigation visualization is shown in \cref{visual_nav}. From the \cref{RL} and \cref{NSRL}, it is evident that NSRL has a wider motion range in terms of the length on the x-axis, breadth on the y-axis, and height on the z-axis compared to RL. This indicates that under the same conditions, NSRL, which benefits from rule enhancement, effectively improves the global navigation capability of the DRL agent. Indeed, NSRL agent possesses a more diverse range of behaviors. This diversity allows the NSRL agent to explore a wider range of areas in the environment, enabling it to gather more valuable experiences. This result assists us in designing more interactive agents that can navigate the entire map more effectively.

\begin{figure}
\vskip 0.1in
    \centering
    \includegraphics[width=\linewidth]{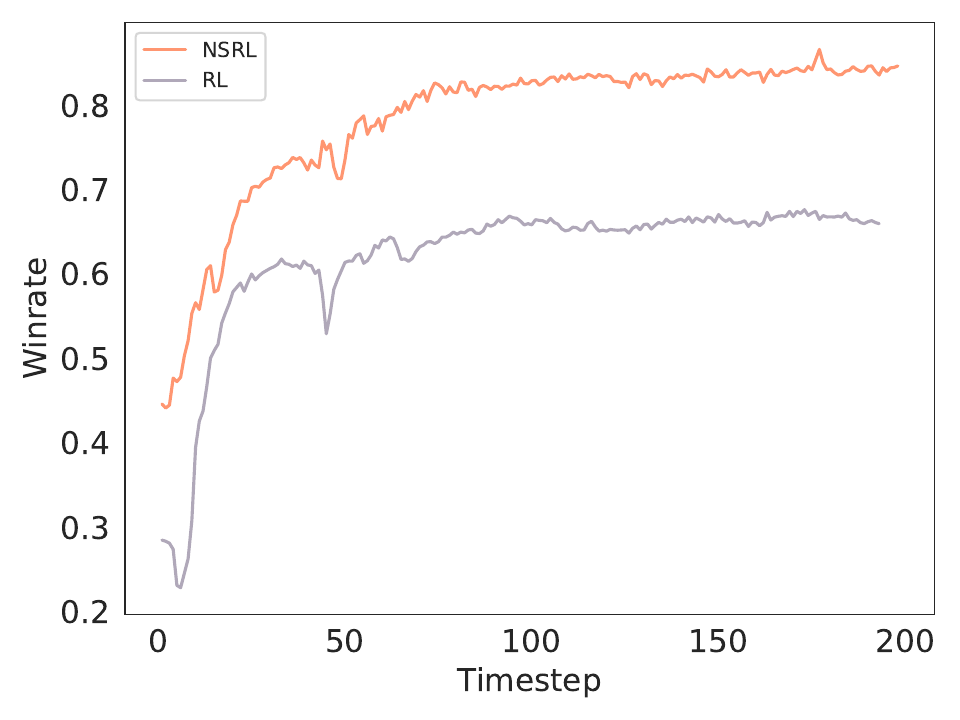}
    \caption{The trend of win rates verses BT changes during the training process for the RL agent and NSRL agent.}
    \label{fig:winrate}
\vskip -0.2in
\end{figure}

\textbf{Competence of NSRL agent.} To demonstrate the competence of NSRL compared to pure RL, both agents were trained using identical resources and opponents (based on expert prior behavior trees) throughout the training process. We extracted the win rate changes of the NSRL agent and RL agent at 200 timesteps and plotted them in \cref{fig:winrate}. At the same timestep, the win rate of NSRL was consistently around 30\% higher on average compared to the RL agent. This highlights the superior performance of NSRL in adapting to expert prior behavior trees. Furthermore, it emphasizes the improved training efficiency of NSRL, as it is capable of producing more competent agents at the same timestep.

\begin{figure*}[ht]
\vskip 0.1in
	\centering
    \subfigure[Bullet Distribution of Human Players] {\includegraphics[width=.3\textwidth]{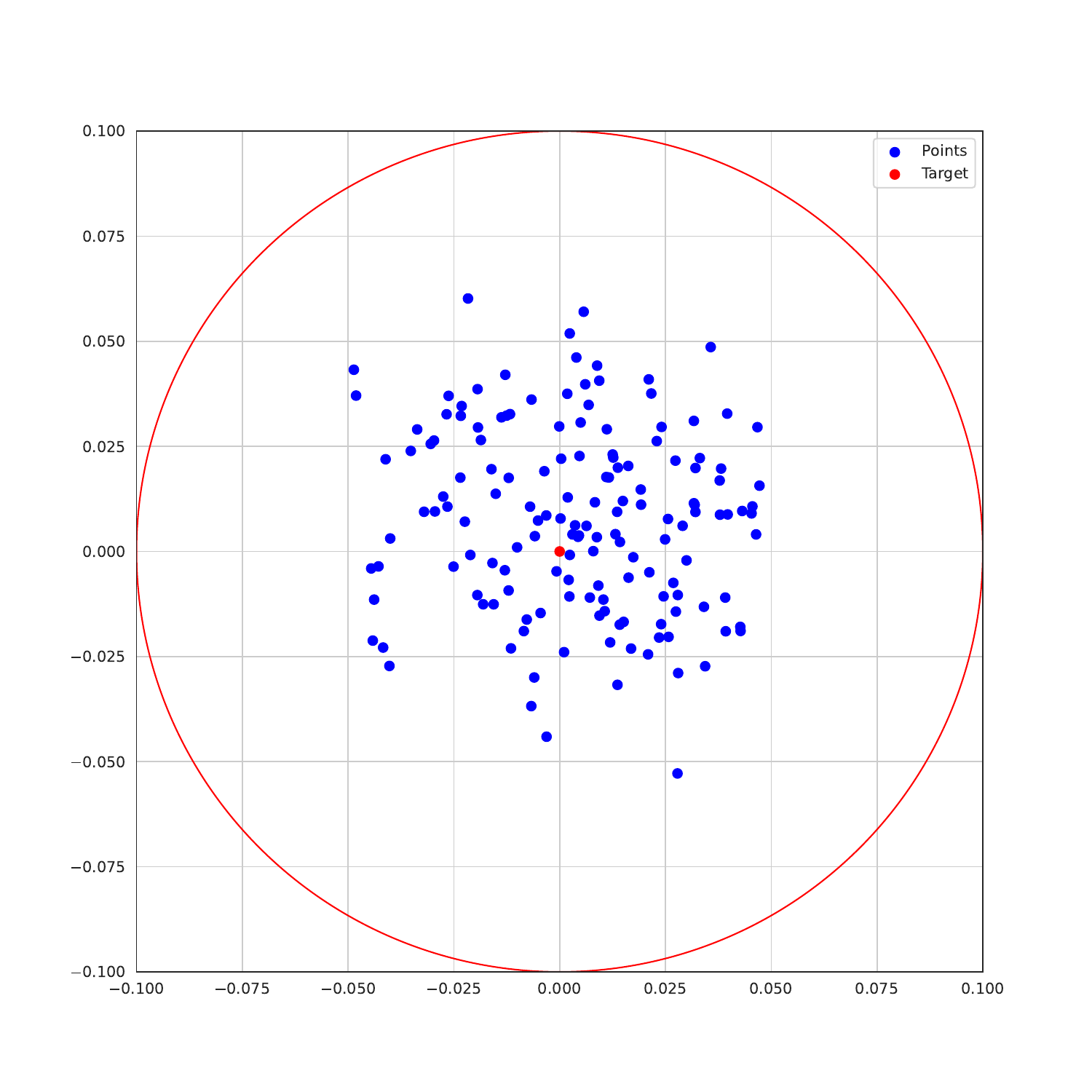}\label{b_human}}
	\subfigure[Bullet Distribution of NSRL agent] {\includegraphics[width=.3\textwidth]{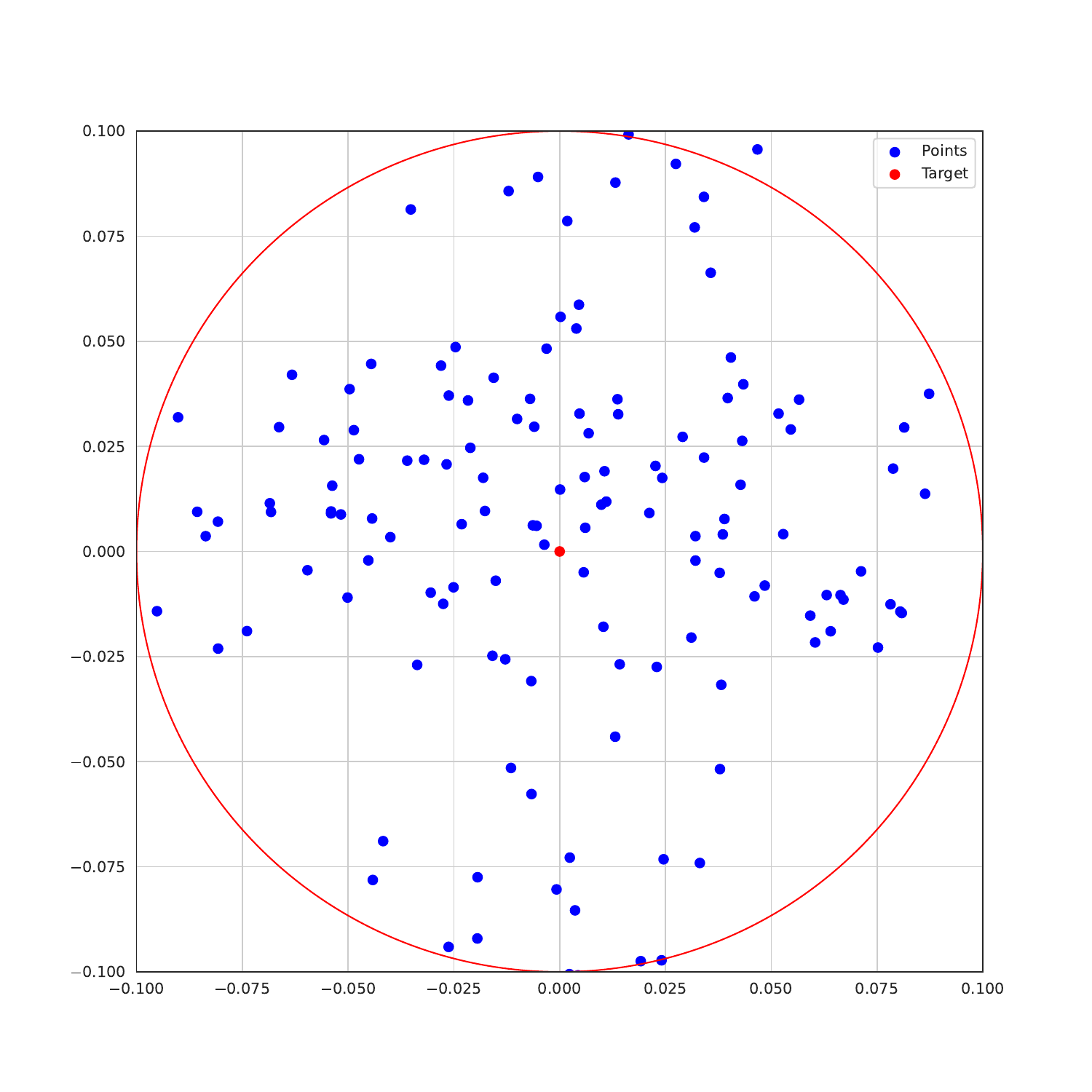}\label{b_NSRL}}
	\subfigure[Bullet Distribution of RL agent] {\includegraphics[width=.3\textwidth]{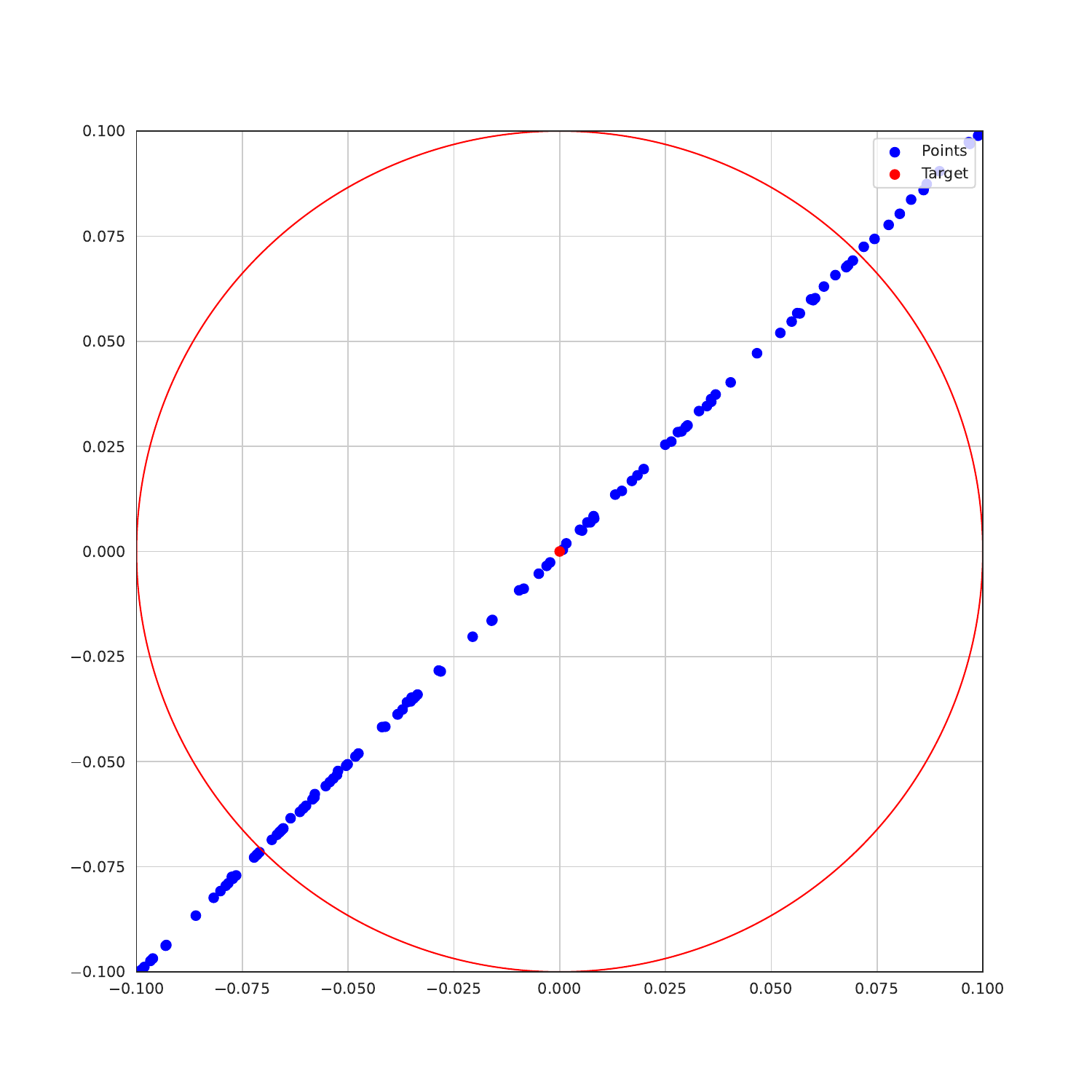}\label{b_RL}}
	\caption{Bullet Distribution of Human Players, NSRL agent, and RL agent.}
	\label{fig:bullet}
\vskip -0.2in
\end{figure*}

\textbf{Shooting of NSRL.} Due to the prevalent use of realistic firearm recoil and ballistic effects in modern 3D FPS games, it is crucial for DRL agents to exhibit human-like shooting behavior that does not confuse players. To analyze this, we designed a specific experimental scenario where the NSRL and RL agents were tasked with shooting at fixed target locations. We collected the coordinates of bullet impact patterns near the target locations and integrated them into \cref{fig:bullet}. From the \cref{b_human}, it can be observed that when human players shoot, their bullets are dispersed around the central target point, which aligns with the logic of realistic firearm recoil and ballistic effects. The result in \cref{b_NSRL} shows the bullet distribution of the NSRL agent exhibits a similar trend to that of humans but with a higher degree of dispersion. This may be due to a lack of sufficient understanding of the concept of "spray control" in FPS games within the model. \cref{b_RL} illustrates that pure RL alone cannot effectively control firearms and may exhibit perplexing behaviors.

In summary, NSRL effectively enhances the global navigation capabilities of DRL agents, enabling them to exhibit more diverse strategies and behaviors. NSRL improves training efficiency while maintaining a certain level of competitiveness, making it a necessary asset for deployment in high-level scenarios. Moreover, NSRL's shooting rule augmentation enhances the human-like nature of DRL agents, ensuring that they do not exhibit confusing behaviors that diminish the players' gaming experience.

\section{RELATED WORK}

\subsection{Reinforcement Learning in FPS Games}
In the field of reinforcement learning, FPS is a highly valuable research environment. In FPS games, it is necessary to balance both movement and firing actions under partial observation and achieve the best performance in a competitive environment. Currently, many FPS games are used for research on reinforcement learning algorithms.
Two 1990s games were introduced to the RL environment in the early stage. \citet{kempka2016vizdoom} introduced ViZDoom and \citet{jaderberg2019human} built Quake III Arena as RL environment. There have been many excellent works published based on these environments currently. The DRQN, proposed by \citet{hosu2016playing}, is one of the most influential works in ViZDoom. DRQN is also modularized to allow different models to be independently trained for different phases of the game which substantially outperforms built-in AI agents of the game as well as average humans in deathmatch scenarios. 
\citet{wu2016training} using Actor-Critic curriculum learning to train vision-based agents in VizDoom and win the champion of Track1 in ViZDoom AI Competition 2016. \citet{jaderberg2019human} demonstrate for the first time that an agent can achieve human-level in a popular 3D multiplayer first-person video game, Quake III Arena Capture the Flag (28), using only pixels and game points as input. The result is achieved by a novel two-tier optimization process in which a population of independent RL agents is trained concurrently from thousands of parallel matches with agents playing in teams together and against each other on randomly generated environments. 

Besides ViZDoom and Quake III Arena, \citet{pearce2022counter} learn an agent to play deathmatch in CS: GO with adopting behavioral cloning. Their method shows reasonably good performance and high data efficiency, a new way to apply RL in FPS games. \citet{chen2022wild} develop WILD-SCAV, a powerful and extensible environment based on a 3D open-world FPS game which is an environment with greater diversity and complexity. They want to bridge the gap that the existing environment is hardly extensible to more complicated problems. zhao at al propose a rule-enhanced deep reinforcement learning algorithm for three tasks. And their agent has better performance over multiple rule-based and RL-based agents.

\section{Conclusion}
In this paper, we propose a full-map interactive agent, PMCA, for Arena Breakout that addresses the challenges of global navigation and realistic ballistic behavior inherent in applying DRL to modern FPS games. To address these challenges, we propose a novel method that combines Navmesh and NSRL. The shooting-rule constraint is used to address the human-like behavior problem, while the Navmesh enhances global movement capabilities. Through experiments, we demonstrate the superiority of NSRL. This agent has been deployed to Arena Breakout since early 2024, marking an important milestone in the practical application of reinforcement learning in gaming. In our future work, we will further investigate the agent's ability to explore large-scale 3D environments by adopting occupancy maps and different computer vision techniques.

\bibliographystyle{unsrtnat}
\bibliography{mybib}

\end{document}